\pdfoutput=1

\documentclass[11pt]{article}

\usepackage{acl}

\usepackage{times}
\usepackage{latexsym}  

\usepackage[T1]{fontenc}

\usepackage[utf8]{inputenc}

\usepackage{microtype}

\usepackage{inconsolata}

\usepackage{graphicx}

\usepackage{amsmath,amsfonts}
\usepackage{algorithmic}
\usepackage{array}
\usepackage[caption=false,font=normalsize,labelfont=sf,textfont=sf]{subfig}
\usepackage{textcomp}
\usepackage{stfloats}
\usepackage{url}
\usepackage{verbatim}
\usepackage{graphicx}
\usepackage{tabularx}
\usepackage{multirow}
\usepackage{hyperref}
\usepackage{booktabs}
\usepackage{multirow}
\usepackage{xcolor}

\usepackage{eso-pic}
\usepackage{xcolor}

\usepackage{eso-pic}
\usepackage{xcolor}

\AddToShipoutPictureFG*{%
  \put(\LenToUnit{\paperwidth/2},8){%
    \makebox(0,0){\fontsize{7}{8}\selectfont\textcolor{gray}{\textbf{CAUTION:} This paper may contain harmful content.}}%
  }%
}

\title{On VLMs for Diverse Tasks in Multimodal Meme Classification}

\author{
Deepesh Gavit, Debajyoti Mazumder,  Samiran Das,  Jasabanta Patro \\
  Department of Data Science and Engineering \\
  Indian Institute of Science Education and Research, Bhopal, India \\
  \texttt{\{gavit20, debajyoti22, samiran, jpatro\}@iiserb.ac.in}\\
}

\begin{document}
\maketitle
\begin{abstract}

In this paper, we present a comprehensive and systematic analysis of vision-language models (VLMs) for disparate meme classification tasks. We introduced a novel approach that generates a VLM-based understanding of meme images and fine-tunes the LLMs on textual understanding of the embedded meme text for improving the performance. Our contributions are threefold: (1) Benchmarking VLMs with diverse prompting strategies purposely to each sub-task; (2) Evaluating LoRA fine-tuning across all VLM components to assess performance gains; and (3) Proposing a novel approach where detailed meme interpretations generated by VLMs are used to train smaller language models (LLMs), significantly improving classification. The strategy of combining VLMs with LLMs improved the baseline performance by 8.34\%, 3.52\% and 26.24\% for sarcasm, offensive and sentiment classification, respectively. Our results reveal the strengths and limitations of VLMs and present a novel strategy for meme understanding.

\end{abstract}
\section{Introduction}

Multi-modal memes have gained prominence \cite{petrova2021meme} due to their eloquent and powerful way to convey complex, subtle messages \cite{das2023rising}. The widespread use of memes propagating hatred \cite{gelber2016evidencing}, abuse, misogyny, and propaganda is a serious cause of concern \cite{bhattacharya2019social}. AI models can safeguard digital platforms \cite{hee2024recent} by comprehending memes. The NLP community has been trying to address this issue for years. This is evident from several publications and datasets published in recent years (See Details in  Appendix (Section~\ref{relatedwork})). 

These prevalent works introduced diverse datasets, presented efficient multi-modal deep learning (DL) models, and explored the capabilities of LLMs and VLMs \cite{afridi2021multimodal}. We mentioned the details of these works in the Appendix. While the DL models used for the task are not generalizable \cite{shah2024memeclip}, LLMs are incapable of processing visual information \cite{cai2025investigation}. The performance of the VLMs used in this task depends on their contextual understanding \cite{xing2024survey}. Therefore, the existing methods are yet to attain satisfactory high efficacy necessary for the real-world deployment of these models. Since the image and the text in the memes are either complementary or indirectly linked, both modalities are essential for the task \cite{zhong2024multimodal}. Besides, accurate meme understanding necessitates relating the information to the contextual understanding and the cultural backdrop \cite{yus2019multimodality}. Due to these predicaments, accurate meme classification is a challenging task for the AI models \cite{jha2024memeguard}.

\begin{figure*} 
    \centering
    \includegraphics[width=0.6\linewidth]{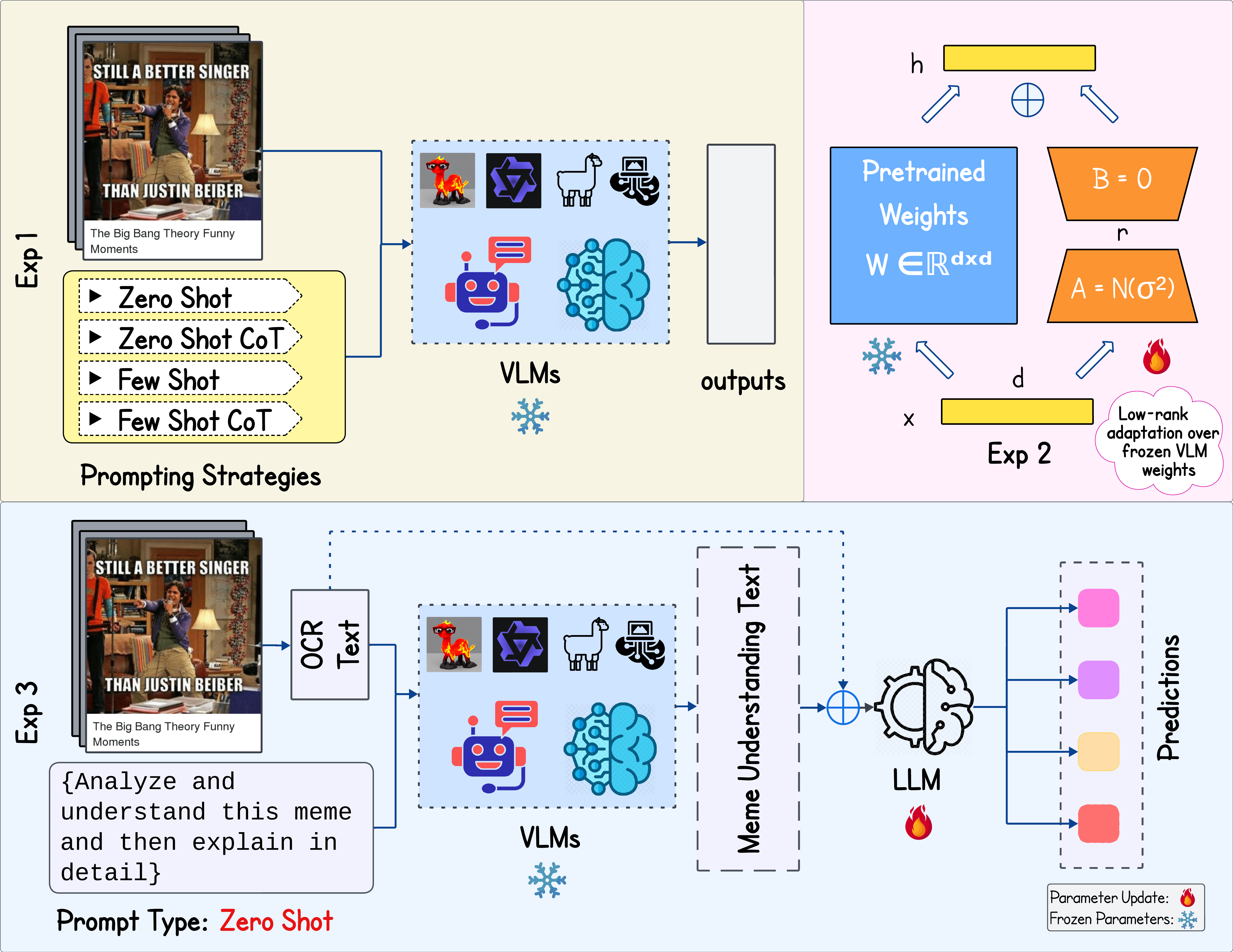} 
    \caption{Schematic diagram of the three strategies mentioned in the experiment section.}
    \label{fig:flowchart_exp3}
\end{figure*}

We utilized widely used open-sourced VLMs in different settings as  illustrated in Fig. \ref{fig:flowchart_exp3} to address the following research questions- 
\begin{itemize}
    \item \textbf{R1:} Are the pre-trained VLMs with diverse types of prompting powerful enough to understand memes?
    \item \textbf{R2:} Does fine-tuning VLMs using adapters improve the performance? 
    \item \textbf{R3:} Can we combine pre-trained VLMs contextual understanding to train the LLMs for better classification?
\end{itemize}

The important contributions of the work are:
\begin{itemize}
    \item \textbf{Benchmarking VLM Meme Classification:} \textit{We benchmark the performance of various VLMs using diverse prompting strategies.}
    \item \textbf{Evaluating LoRA Fine-Tuning:}\textit{ We systematically explore diverse prompting strategies such as Zero Shot (ZS), Zero Shot Chain-of-Thought (ZSC), Few Shot (FS), Few Shot Chain-of-Thought (FSC), and parameter-efficient LoRA fine-tuning approach.} 
    \item \textbf{Novel Meme Understanding Approach:} \textit{We introduce an innovative approach called Combining VLM Explanation to Fine-tune LLMs (CoVExFiL). In this approach, we use VLM-generated meme interpretations from multiple prompting techniques to train LLMs. The proposed approach with a three-step CoT prompting produces the best results.}
\end{itemize}

Some of the key findings from this paper are as follows-
\begin{itemize}
    \item Fine-tuning VLMs using LoRA adapter was relatively less effective for our task, because it only updates a limited number of parameters and hence do not learn the cultural context. On the other hand, the Qwen model surpassed the baseline performance for sentiment and sarcasm classification under FS prompting by 16.71\%, and 0.39\%, respectively.

    \item The proposed CoVExFiL strategy yields noticeable performance gains in tasks such as SR and OF, with a significant 26.10\% improvement in SN, outperforming SOTA. This is attributed to its strong meme understanding, and error analysis further confirms the effectiveness of training on generated explanations.

\end{itemize}

\section{Datasets and Experimental Setup}

We used two popular datasets, i.e., Memotion by ~\citet{sharma2020semeval} and multimedia automatic misogyny identification (MAMI) by ~\citet{fersini2022semeval} in our experiments. While memotion categorizes the memes in five emotion categories (humour (HM), sarcasm (SR), offensiveness (OF), sentiment (SN), and motivational (MV)), MAMI labels whether a meme is misogynistic (MG) and categorizes the type of misogyny (as detailed in Appendix ~\ref{datasets_}).  We use open-source vision-language models; LLaVA (LV), Qwen (QW), LLaMA (LM), and InstructBLIP (IB) for our experiments. For context, we also include results from deep-learning models (SOTA) trained individually on each dataset for each task as reference benchmarks. Since these models are tailor-made to specific tasks, they are used only for comparison not as baselines. Details of these models can be found in the Appendix ~\ref{model_config}.

\section{Experiments}

In this section, we have reported the details of various experiments conducted as a part of this study. Particularly, we conducted three types of experiments, each based on a unique research philosophy. While the first experiment prompts the VLMs using various prompting strategies, the second experiment finetunes the LoRA adapter associated with the VLMs. Lastly, in the third experiment, we finetuned the LLMs by giving VLM-generated meme understanding as input. We reported the details of the individual experiment in subsections.

\subsection{Experiment-1 (Exp1): Prompting VLMs using various methods}

In this experiment, we evaluated various prompting strategies such as \textit{ZS}, \textit{ZSC}, \textit{FS}, and \textit{FSC} to classify memes. While in the ZS setting we prompted VLMs to classify the memes, in ZSC \cite{wei2022chain},\cite{xu-etal-2024-exploring} we additionally asked the VLMs to provide detailed step-by-step reasoning. In the FS version, we followed similar steps; however, in addition, we included some example input for context. The FSC setting combined the FS approach with CoT-based reasoning. Prompting has been shown to influence model outputs across a range of visual reasoning tasks ~\cite{zhou2024image}. More details are provided in the Appendix~\ref{exp_1}.

\subsection{Experiment-2: Fine-tuning VLMs using LoRA adapter} 

In this experiment, we explored a substitute option of employing LoRA (Low-Rank Adaptation)~\cite{hu2022lora}, a parameter-efficient approach for fine-tuning VLMs. Instead of updating all model parameters, LoRA trains lightweight adapter layers, enabling VLMs to efficiently adapt to the meme classification task. Further details of the LoRA method are provided in the Appendix~\ref{exp_2}. Since the performance of the pre-trained VLMs depends on the training data, they perform well only if the model is trained on similar data. 




\subsection{Experiment-3: Combining VLM Explanation to Fine-tune LLMs} 
 In this experiment, CoVExFiL, we incorporated a novel two-step approach where, i) first, we generated textual understanding of the memes by prompting VLMs, and subsequently, ii) we fine-tuned the LLMs (details in Appendix~\ref{llms_config}) using the scene understanding generated by the VLMs. Our objective was to investigate whether VLM-generated understanding explanations can help improve the LLMs to classify the memes. In this experiment, we fine-tuned the LLMs, which is relatively easier \cite{chung2024scaling} instead of fine-tuning the VLMs.

\renewcommand{\arraystretch}{1.2} 
\setlength{\tabcolsep}{4pt} 

\begin{table}[h!]
\centering
\setkeys{Gin}{keepaspectratio}
\resizebox{1\columnwidth}{!}{%
\begin{tabular}{|l|c|c|c|c|c|c|c|c|c|}
\hline
\textbf{MDL} & \textbf{PM} & \multicolumn{5}{c|}{\textbf{Memotion}} & \multicolumn{2}{c|}{\textbf{MAMI}} & \textbf{Avg.} \\ \cline{3-9}
             &             & \textbf{HM} & \textbf{SR} & \textbf{OF} & \textbf{SN} & \textbf{MV} & \textbf{MG} & \textbf{MGT} &              \\ \hline

\multirow{4}{*}{LV} 
& ZS  & 27.43 & 13.82 & 11.57 & 24.73 & 25.67 & 57.31 & 14.65 & 25.03 \\ 
& ZSC & 29.83 & 12.76 & 24.55 & 32.35 & 47.33 & 68.38 & 31.06 & 35.18 \\ 
& FS  & 25.94 & \textbf{\textcolor{blue}{32.98}} & 12.72 & 16.57 & 51.42 & 75.46 & 34.93 & 35.72 \\ 
& FSC & 28.30 & 18.92 & 23.44 & 27.91 & 55.96 & \textbf{\textcolor{blue}{77.65}} & 28.77 & 37.28 \\ \hline

\multirow{4}{*}{QW} 
& ZS  & 25.97 & 14.82 & 19.89 & 35.49 & 51.58 & 49.12 & 13.24 & 30.02 \\ 
& ZSC & 27.29 & 12.79 & 24.88 & 36.09 & \textbf{\textcolor{blue}{57.62}} & 71.77 & 25.55 & 36.57 \\ 
& FS  & 27.57 & 10.23 & \textbf{\textcolor{blue}{29.41}} & \textbf{\textcolor{blue}{44.70}} & 53.06 & 70.87 & 25.74 & 37.37 \\ 
& FSC & \textbf{\textcolor{blue}{31.34}} & 22.18 & 27.72 & 44.19 & 52.69 & 73.31 & \textbf{\textcolor{blue}{38.92}} & 41.48 \\ \hline

\multirow{2}{*}{LM} 
& ZS  & 24.76 & 18.23 & 20.79 & 20.32 & 19.34 & 51.65 & 12.37 & 23.92 \\ 
& ZSC & 29.91 & 17.85 & 22.68 & 41.48 & 49.80 & 60.76 & 27.79 & 35.75 \\ \hline

\multirow{2}{*}{IB} 
& ZS  & 22.85 & 19.47 & 13.69 & 11.25 & 18.12 & 45.47 & 11.18 & 20.29 \\ 
& ZSC & 23.85 & 28.64 & 22.69 & 20.25 & 45.12 & 49.47 & 20.18 & 30.03 \\ \hline \hline

\multicolumn{2}{|l|}{\textbf{Avg.}} 
& 27.09 & 18.56 & 21.17 & 29.61 & 43.98 & 62.60 & 23.70 &  \\

\multicolumn{2}{|l|}{\textbf{Std.}} 
& $\pm$2.56 & $\pm$6.72 & $\pm$5.78 & $\pm$11.17 & $\pm$14.33 & $\pm$11.64 & $\pm$9.29 & \\ \hline

\multicolumn{2}{|l|}{\textbf{SOTA \(\triangle\)}} 
& 49.09 & 32.85 & 34.38 & 38.30 & 59.28 & 87.40 & 73.14 &  \\ \hline
\end{tabular}
}
\caption{Performance of VLMs via prompting approaches. The highest F1 score for each task is in \textbf{\textcolor{blue}{blue}} (column-wise).}
\label{tab:vlm_evaluation}
\end{table}

\begin{table}[htbp]
\centering
\setkeys{Gin}{keepaspectratio}
\resizebox{0.95\columnwidth}{!}{%
\begin{tabular}{|l|c|c|c|c|c|c|c|c|}
\hline
\textbf{MDL} & \multicolumn{5}{c|}{\textbf{Memotion}} & \multicolumn{2}{c|}{\textbf{MAMI}} & \textbf{Avg.} \\ \cline{2-8}
             & \textbf{HM} & \textbf{SR} & \textbf{OF} & \textbf{SN} & \textbf{MV} & \textbf{MG} & \textbf{MGT} &              \\ \hline
LV & 30.21 & 17.58 & 11.57 & 29.83 & 49.16 & \textbf{64.16} & 24.53 & 32.43 \\ \hline
QW & \textbf{\textcolor{blue}{36.35}} & 20.97 & \textbf{\textcolor{blue}{29.89}} & 30.34 & \textbf{\textcolor{blue}{51.58}} & 61.33 & \textbf{\textcolor{blue}{27.83}} & 36.89 \\ \hline
LM & 33.95 & \textbf{\textcolor{blue}{23.17}} & 26.34 & \textbf{\textcolor{blue}{31.72}} & 49.33 & 58.98 & 22.13 & 35.08 \\ \hline
IB & 23.03 & 15.98 & 18.98 & 28.61 & 35.19 & 47.33 & 19.83 & 26.99 \\ \hline \hline
\textbf{Avg.} & 30.89 & 19.43 & 21.70 & 30.13 & 46.32 & 57.95 & 23.58 & \\ 
\textbf{Std.} & $\pm$4.86 & $\pm$2.84 & $\pm$7.52 & $\pm$1.28 & $\pm$6.66 & $\pm$6.50 & $\pm$3.37 & \\ \hline 
\textbf{SOTA \(\triangle\)} & 49.09 & 32.85 & 34.38 & 38.30 & 59.28 & 87.40 & 73.14 & \\ \hline
\end{tabular}
}
\caption{LoRA Adapter Fine-Tuning (Experiment 2). The highest F1 score for each task is shown in \textbf{\textcolor{blue}{blue}} (column-wise).}
\label{tab:lora_finetuning}
\end{table}

\section{Results and Discussion} 

We analyze the diverse memes using three different strategies, each based on a distinct philosophy.  We quantify the performance using \textbf{average weighted F1 (AWF1) score} to counter the class imbalance issue. We compare results task-wise and model-wise, highlight key trends, and show how each strategy enhances meme classification performance.

\subsection{Observations for Experiment 1}

Table~\ref{tab:vlm_evaluation} demonstrates that FSC prompting improves performance in all tasks, with QW achieving the highest AWF1 \textbf{41.48}. This result highlights the importance of structured reasoning and its effectiveness in enhancing multimodal understanding. Task-wise, MG shows the best performance (\textbf{77.65}), likely due to the presence of explicit cues. In contrast, tasks like SR and OF resulted in lower AWF1 (\textbf {18.56}–\textbf{29.61}) on average, reflecting challenges in contextual and complexity. Model-wise, QW and LV respond best to prompting strategies, with LV attaining \textbf{49\%} gain from ZS to FSC. In contrast, IB and LM perform poorly under ZS (\textbf{20.29}, \textbf{23.92}), indicating limited reasoning without guidance. Although prompting lags behind SOTA, three-step CoT shows promise. Notably, in tasks such as MG and MV, the performance gains remain modest.

\begin{table*}
\centering
\setkeys{Gin}{keepaspectratio}
\resizebox{\textwidth}{!}{%
\begin{tabular}{l l r r r r r r r r r r r r r r r r r r r r r r}
\hline
             & \multicolumn{18}{c}{\textbf{Memotion}}                        & \multicolumn{2}{l}{\textbf{MAMI}} &  \\ \midrule \hline
\textbf{MDL} & \textbf{PM} & \multicolumn{3}{c}{\textbf{HM}} & \multicolumn{3}{c}{\textbf{SR}} & \multicolumn{3}{c}{\textbf{OF}} & \multicolumn{3}{c}{\textbf{SN}} & \multicolumn{3}{c}{\textbf{MV}} & \multicolumn{3}{c}{\textbf{MG}} & \multicolumn{3}{c}{\textbf{MGT}} & \textbf{Avg.} \\ \hline
 &  & BR & RB & XL & BR & RB & XL & BR & RB & XL & BR & RB & XL & BR & RB & XL & BR & RB & XL & BR & RB & XL &  \\ \hline

\multirow{4}{*}{LV} 
& ZS    & 26.57 & 16.24 & 28.95 & 26.23 & 15.97 & 19.28 & 26.36 & 20.52 & 22.16 & 26.53 & 25.56 & 31.79 & 51.96 & 16.75 & 54.03 & 66.95 & 67.80 & 67.89 & 53.07 & 43.29 & 44.96 & 35.85 \\ 
& ZSC & 32.01 & 31.37 & 29.37 & 32.70 & 34.93 & \textbf{\textcolor{blue}{35.59}} & 35.04 & 31.97 & 32.14 & \textbf{48.22} & 44.47 & 47.87 & \textbf{\textcolor{blue}{58.67}} & 46.75 & 55.20 & 68.52 & 66.51 & 68.51 & 54.33 & 51.97 & 57.10  & 46.12 \\ 
& FS & 35.90 & \textbf{32.95} & 34.59 & 30.54 & 34.93 & 33.01 & 34.10 & 30.26 & 34.35 & 47.01 & 43.18 & 47.32 & 54.68 & 53.07 & 56.75 & 56.94 & 60.14 & 47.68 & 52.82 & 48.53 & 50.52  & 43.77 \\ 
& FSC & 36.59 & 31.15 & 32.85 & \textbf{34.94} & \textbf{35.08} & 35.45 & 31.42 & 30.53 & 32.16 & 47.08 & 45.39 & 43.83 & 48.20 & 46.75 & 49.98 & 61.08 & 60.09 & 55.01 & 50.01 & 52.07 & 51.08 & 39.16 \\ \hline

\multirow{4}{*}{QW} 
& ZS    & 27.42 & 11.37 & 21.47 & 30.41 & 18.56 & 25.24 & 29.17 & 21.56 & 25.46 & 36.02 & 25.39 & 27.32 & 55.07 & 37.89 & 46.58 & 71.70 & 67.09 & 71.38 & 52.84 & 56.12 & 56.56 & 38.79 \\ 
& ZSC & 31.66 & 32.44 & \textbf{\textcolor{blue}{38.46}} & 24.53 & 26.67 & 35.33 & 32.98 & 28.94 & 24.52 & 43.66 & \textbf{47.51} & \textbf{\textcolor{blue}{48.31}} & 57.02 & 36.75 & 55.29 & 71.59 & \textbf{75.08} & \textbf{\textcolor{blue}{79.45}} & \textbf{\textcolor{blue}{58.47}} & \textbf{56.33} & 57.56 & 45.84  \\ 
& FS & \textbf{38.29} & 31.43 & 35.92 & 32.26 & 32.02 & 31.81 & \textbf{\textcolor{blue}{35.61}} & 32.53 & \textbf{34.77} & 46.57 & 44.47 & 39.61 & 55.01 & 53.21 & \textbf{56.95} & 33.51 & 36.43 & 33.52 & 56.69 & 53.83 & 56.99 & 41.50 \\ 
& FSC & 35.13 & 31.89 & 33.46 & 33.89 & 31.29 & 32.52 & 34.76 & 30.64 & 31.48 & 41.58 & 42.17 & 40.19 & 55.63 & \textbf{56.05} & 56.49 & 64.85 & 68.71 & 71.46 & 51.62 & 54.02 & \textbf{57.83} & 45.51 \\ \hline

\multirow{2}{*}{LM} 
& ZS    & 29.02 & 21.37 & 30.41 & 28.95 & 17.78 & 25.67 & 28.64 & 30.35 & 28.24 & 37.15 & 25.39 & 33.61 & 46.37 & 36.75 & 48.24 & \textbf{71.73} & 71.50 & 76.01 & 55.38 & 56.01 & 56.06 & 40.70 \\ 
& ZSC & 30.18 & 31.37 & 32.45 & 28.93 & 28.91 & 29.25 & 34.77 & \textbf{33.20} & 26.97 & 45.63 & 35.23 & 45.59 & 56.69 & 53.21 & 53.97 & 68.26 & 70.85 & 70.40 & 56.23 & 56.12 & 56.33 & 44.98 \\ \hline

\multirow{2}{*}{IB} 
& ZS    &25.15 & 16.24 & 21.47 & 24.62 & 18.94 & 29.34 & 23.20 & 22.20 & 17.87 & 29.63 & 30.78 & 25.42 & 49.35 & 36.75 & 42.90 & 60.15 & 62.81 & 60.86 & 52.34 & 53.82 & 52.10 & 36.00 \\ 
& ZSC & 32.25 & 30.10 & 28.37 & 31.02 & 28.89 & 30.03 & 26.69 & 28.98 & 29.28 & 43.46 & 33.83 & 35.39 & 55.21 & 55.93 & 38.51 & 64.39 & 63.09 & 64.81 & 54.44 & 54.38 & 55.65 & 42.13  \\ \hline

\multirow{1}{*}{Avg.} 
   & & 31.68 & 26.49 & 30.65 & 31.17 & 27.00 & 30.21 & 31.06 & 28.47 & 28.28 & 41.06 & 36.95 & 38.85 & 53.66 & 44.16 & 51.24 & 63.31 & 64.18 & 63.92 & 54.02 & 53.06 & 54.69 \\

\multirow{1}{*}{Std.} 
& & $\pm$3.95 & $\pm$6.28 & $\pm$3.20 & $\pm$3.79 & $\pm$7.13 & $\pm$5.82 & $\pm$4.68 & $\pm$5.49 & $\pm$6.25 & $\pm$7.84 & $\pm$7.29 & $\pm$7.68 & $\pm$4.56 & $\pm$7.35 & $\pm$6.01 & $\pm$5.18 & $\pm$5.84 & $\pm$6.97 & $\pm$2.79 & $\pm$3.27 & $\pm$2.94 \\  
 
\multirow{1}{*}{SOTA} 
 &  &  & 49.09 &  &  & 32.85 &  &  & 34.38 &  &  & 38.30 &  &  & 59.28 &  &  & 87.40 &  &  & 73.14 \\ \hline
\end{tabular}
}
\caption{Finetuning LLMs on VLM-generated explanations (Exp 3). Notation: The highest F1 score of each LLM is in \textbf{bold} (column-wise) and the highest F1 score in each task is in \textcolor{blue}{blue}. LM and IB does not support FS, and FSC prompting.}
\label{tab:vlm_underst_llm_finetune_1}
\end{table*}
\subsection{Observations for Experiment 2}
Table ~\ref{tab:lora_finetuning} reports the efficacy after using LoRA fine-tuning, which affirms that this method does not lead to noticeable improvement. Tasks such as HM, SR, OF, SN, MV, MG, and MGT report relative improvement of 14.01$\%$, 4.69, 2.50, 1.76, 5.32, -7.43, -0.51, respectively. MV and MG detection tasks attained relatively higher scores, while SR and OF remained challenging. Compared to Exp 1, even if LoRA underperforms in terms of best overall model scores (e.g., QW-FSC: \textbf{41.48} in Exp 1 vs. QW-LoRA: \textbf{36.89}).  While LoRA finetuning slightly improves task-wise averages, prompting-based methods, especially CoT, consistently obtain better model performance. Overall, QW-FSC achieves significantly better results with prompting (e.g., 41.48 vs. 36.89), showing that prompting strategies handle complex meme understanding more effectively than LoRA.

\subsection{Observations for Experiment 3}

Table~\ref{tab:vlm_underst_llm_finetune_1} shows that CoVExFiL significantly outperformed prior experiments, achieving relative gains of 23.10\% (HM), 7.92\% (SR), 21.01\% (OF), 8.07\% (SN), 1.86\% (MT) and 2.32\% (MG) over the best prompting scores (Exp 1). COVExFiL reports \textbf{26.14\%} improvement over the SOTA for SN, while SR (8.34\%) and OF (3.58\%) saw only moderate gains—SR often masks negative sentiment behind positive wording. 

We also included a detailed error analysis in Appendix \ref{error_analysis}. The error analysis revealed persistent over-reliance on surface features, missed implicit offensiveness and visual irony, and strict metrics that penalize near-miss predictions, tempering SR and OF improvements. Nevertheless, we find that higher F1 scores align with CoT-based understandings, with QW achieving the highest BERT Scores under CoT prompting and LV with ZSC prompting attained an average F1 of 46.12\%, representing an 11.2\% increase over Exp 1 (QW-FSC: 41.48\%) and 25.0\% over Exp 2 (QW-LoRA: 36.89\%). These results demonstrate that VLM explanation–based fine-tuning outperforms both direct prompting and adapter tuning, and that three-step CoT prompting (ZSC, FSC) produces more structured explanations that enhance downstream LLM fine-tuning, collectively driving significant classification accuracy gains.

\section{Conclusion}

In this treatise, we presented a systematic, in-depth study on utilizing VLMs in diverse strategies for accurate meme classification. The analysis affirmed that pre-trained VLMs generally perform well in the presence of explicit cues in tasks like SN and MG. However, this approach cannot accurately comprehend nuanced content like SR or OF. The prompting strategies, particularly CoT, improved the reasoning and classification accuracy. However, LoRA-based fine-tuning proved to be less effective, mostly because LoRA alters a relatively small number of parameters from very few selective layers. Our proposed CoVExFiL approach, which integrates VLMs and LLMs, performs well in multiple tasks. These findings underscore the effectiveness of prompting and distillation for improving meme understanding.

Our analysis shows that VLMs grasp meme context well when the content contains clear, straightforward clues. However, they struggle with hidden meanings, particularly in OF and SR tasks. The CoT provides moderate performance gains and helps bridge this gap to some extent. 



\section{Limitations}

The primary limitations of our work are described below: First, we used only publicly available VLMs and did not include larger or closed-source models that might deliver stronger reasoning and different insights. Second, we explored four distinct prompting strategies, yet we did not cover the full spectrum of possible prompt variations, which may affect performance. Third, we used LoRA-based fine-tuning instead of complete model fine-tuning due to high computational costs, possibly limiting adaptability and effectiveness. Fourth, we treated each meme independently and relied on publicly available meme explanations, which may lack external context, such as historical events or social media trends, and reflect annotator biases or miss cultural nuances. Addressing this requires more straightforward guidelines, culturally aligned annotators, and familiarity-based filtering. Addressing these issues through broader model selection, richer evaluation, full fine-tuning, external knowledge integration, and multilingual support offers a clear path for future work.

\bibliography{custom}

\appendix

\section{Detailed Related Work}
\label{relatedwork}

Over the past few years, multi-modal meme analysis has emerged as a prominent research area in NLP and multi-modal learning\cite{nguyen2024computational}. These works generally focused on understanding the interplay of visual and textual information in memes for addressing a wide range of tasks, such as sentiment and emotion detection, humor and sarcasm recognition, identification of misogynistic or offensive content, figurative language interpretation, and bias assessment in model predictions \cite{afridi2021multimodal}. Several specialized corpora have been released to capture the rich variety of figurative, humorous, and harmful content in memes. Xu et al. \cite{xu2022met} introduced MET-meme, a collection targeted at metaphorical memes, demonstrating that models struggle when literal text masks non-literal intent. Liu et al. \cite{liu2022figmemes} released FigMemes, covering sarcasm, irony, hyperbole, and metaphor, and showed that integrating social-political context via the MSDBert architecture proposed in the work substantially improves detection of figurative language. Suryawanshi et al.\cite{suryawanshi2020multimodal} proposed MultiOFF, annotated for fine-grained offensiveness and satire, highlighting the role of cultural and linguistic factors in perceiving offense. Fersini et al. \cite{fersini2022semeval} curated the misogyny-focused SemEval-2022 Task 5 benchmark, which goes beyond binary labels to classify different types of gendered hate. Finally, the Memotion and MAMI datasets provide multi-dimensional labels (humor, sentiment, intent), enabling simultaneous evaluation of positive and harmful meme aspects \cite{ramamoorthy2022memotion, afridi2021multimodal}.

Recently, researchers proposed some tailor-made deep learning models to explicitly model both intra- and inter-modal interactions in these datasets. Pan et al. \cite{pan2020modeling} proposed a dual attention framework—separate intra-modal and inter-modal blocks to capture contradictions between text and image, significantly boosting sarcasm detection performance. Liu et al.\cite{liu2022figmemes} MSDBert fuses cross-modal embeddings within a BERT backbone to better capture subtle figurative cues in FigMemes \cite{liu2022figmemes}. Ramamoorthy et al. \cite{ramamoorthy2022memotion} introduced a framework, termed DISARM, to augment standard multi-modal fusion with Named Entity Recognition. The work enables the model to detect harmful content and pinpoint specific targeted entities \cite{ramamoorthy2022memotion}. These studies consistently demonstrate that explicit attention to modality interactions, rather than merely concatenating the modailities, leads to substantial gains across humor, sarcasm, and harm-detection sub-tasks.




Early works like Hendricks et al.\cite{hendricks2018grounding} introduced grounded visual explanations, aligning model justifications with image regions. Jia et al.\cite{jia2017adversarial} showed that explanation generation enhances trust in NLP models, a principle now extended to multi-modal meme understanding. Radford et al.\cite{radford2021learning} demonstrated the ability of CLIP to generate interpretable outputs for memes by aligning visual and textual modalities. Sharma et al.\cite{sharma2023you} proposed \textbf{LUMEN}, a framework for humor classification in memes, combining multi-modal understanding with contextual explanation generation.
VLMs, meanwhile, have revolutionized meme analysis by enabling semantically aligned, multi-modal representations. VisualBERT~\cite{li2019visualbert} and ViLBERT~\cite{lu2019vilbert} introduced unified and dual-stream architectures, respectively, for vision-language fusion. UNITER~\cite{chen2020uniter} and OSCAR~\cite{li2020oscar} enhanced alignment through large-scale pretraining and object tag anchoring. CLIP~\cite{radford2021learning} set a new standard with contrastive learning and zero-shot classification. Models like MiniGPT-4~\cite{zhu2023minigpt} extended these capabilities by incorporating visual grounding and transformer-based vision processing. Additional strategies, such as caption enrichment~\cite{blaier2021caption}, multi-task learning~\cite{lee2022multi}, and external knowledge integration~\cite{pramanick2021momenta}, have been shown to improve meme classification performance and interpretability. \cite{jha2024memeguard} proposed a MemeGuard framework fine-tuned with a dedicated VLM for harmful meme interpretation, applied a multi-modal knowledge-selection module, and then prompts a general-purpose LLM to generate context-aware interventions.

Despite advancements in meme understanding, current deep learning models often fail to generalize beyond the datasets they are trained on, limiting their real-world applicability. Most lack \textbf{interpretability}, offering no insight into their predictions—an issue in sensitive contexts. Additionally, \textbf{robustness} to adversarial content and \textbf{generalization} across cultures remain largely unaddressed. Addressing these gaps is key to building adaptable, trustworthy meme understanding systems.


\begin{table}[h]
\centering
\resizebox{1\columnwidth}{!}{%
\begin{tabular}{|c|c|c|c|c|}
\hline
\textbf{Dataset} & \textbf{Task} & \textbf{Labels}      & \textbf{\# Samples} & \textbf{Length} \\ \hline
\multirow{17}{*}{Memotion} & \multirow{3}{*}{Sentiment (SN)} & Positive & 4,160  & 13.27 \\ 
                          & & Neutral & 2,201 & 12.85 \\ 
                          & & Negative & 631  & 13.57 \\ \cline{2-5}
& \multirow{4}{*}{Humor (HM)} & Not Funny & 1,651 & 13.71 \\ 
                          & & Funny & 2,452 & 12.69 \\ 
                          & & Very Funny & 2,238  & 13.02 \\ 
                          & & Hilarious & 651  & 13.09 \\ \cline{2-5}
& \multirow{4}{*}{Sarcasm (SR)}   & Not Sarcastic  & 1,544  & 13.19 \\ 
                          & & General & 3,507  & 13.02 \\ 
                          & & Twisted Meaning & 1,532  & 13.08 \\ 
                          & & Very Twisted   & 394  & 13.10 \\ \cline{2-5}
& \multirow{4}{*}{Offensive (OF)}& Not Offensive  & 2,713  & 13.31 \\ 
                          & & Slight & 2,592  & 13.00 \\ 
                          & & Very Offensive & 1,466  & 12.74 \\ 
                          & & Hateful Offensive & 221  & 13.21 \\ \cline{2-5}
& \multirow{2}{*}{Motivation (MV)} & Not Motivational & 4,525  & 13.01 \\ 
                          & & Motivational & 2,467 & 13.19 \\ \hline

\multirow{6}{*}{MAMI} & \multirow{2}{*}{Misogyny (MG) } & Misogynous & 5,500 & 16.21 \\ 
                      & & Non-Misogynous & 5,500 & 19.90 \\ \cline{2-5}
& \multirow{4}{*}{Misogyny type (MGT)} & Shaming & 1,420 & 18.54 \\ 
                       & & Stereotype & 3,160 & 18.34 \\ 
                       & & Objectification & 2,550 & 18.93 \\ 
                       & & Violence & 1,106 & 18.26 \\ \hline
\end{tabular}%
}
\caption{Dataset statistics. Here, `Length' denotes the average caption length in words.}
\label{tab:dataset_stat}
\end{table}

\section{Experimental Setup}

\subsection{Datasets Details} \label{datasets_}
This section includes the details of the two datasets, namely, \textbf{Memotion} dataset ~\citet{sharma2020semeval} and Multimedia Automatic Misogyny Identification dataset (\textbf{MAMI} hereafter) ~\citet{fersini2022semeval}, used for extensive experimentation. The dataset statistics are given in Table \ref{tab:dataset_stat}, and dataset characteristics are detailed in the following subsection-

\subsubsection{Memotion Dataset\label{subsec:memotion_dataset}}%
The memotion dataset (\citet{sharma2020semeval}) (publicly available on Kaggle) contains 9,871 multi-modal memes collected from Google images. The dataset creation process involved selecting memes from 52 categories, including political figures and popular cultural references such as Hillary, Trump, Minions, and Baby Godfather. Only memes with embedded English text were retained to ensure linguistic consistency. The memes were annotated using Amazon Mechanical Turk\footnote{\url{https://www.mturk.com/}}, where five annotators independently judged each meme, and a majority voting scheme was used to combine the annotations. The memes were labeled according to five emotion-oriented dimensions: \textit{humor, sarcasm, offensiveness, motivation, and sentiment}. In particular, the details of the classification tasks are as follows-

\begin{itemize}
    \item \textbf{Task A: Sentiment Classification} \\
    Classifies memes into three classes: \textit{Positive}, \textit{Neutral}, and \textit{Negative}.
    \item \textbf{Task B: Emotion Classification} \\
    Labels memes based on one of the following emotions: \textit{humor}, \textit{Sarcasm}, \textit{Offensiveness}, or \textit{Motivation}.
    \item \textbf{Task C: Scales of Semantic Classes} \\
    Categories memes into overlapping emotional scales:
    \begin{itemize}
       \item \textbf{Humor}: not\_funny, funny, very\_funny, hilarious.
        \item \textbf{Sarcasm}: not\_sarcastic, general, twisted\_meaning, very\_twisted.
        \item \textbf{Offensiveness}: not\_offensive, slight, very\_offensive, hateful\_offensive.
        \item \textbf{Motivation}: not\_motivational, motivational.
    \end{itemize}
\end{itemize}

We have visualized some sample memes and their corresponding class labels from the Memotion dataset in Fig. 2. 

\begin{figure*} 
    \centering
    \includegraphics[width=1\textwidth]{images/memotion_examples.pdf} 
    \caption{Data samples from the Memotion dataset. For each meme, the full set of subcategories corresponding to each classification task is listed. The ground-truth label for each task is highlighted in  \textcolor{green}{green}.}
    \label{fig:dataset_samples_memotion_dataset}
\end{figure*}

\subsubsection{MAMI}
The MAMI dataset \citet{fersini2022semeval} (Apache 2.0 licensed) detects misogyny in memes and sub-classifies the type of misogynist content. The memes in this dataset were collected from two sources- social media platforms such as Twitter and Reddit, and meme-sharing websites like 9GAG, Imgur, and KnowYourMeme. These memes were collected by using specific hashtags such as \#girl, \#girlfriend, \#women, \#feminist, threads, and discussions covering feminist debates and other similar events. The dataset was annotated using the crowdsourcing concept. The dataset consists of two types of labels: i) whether the meme is misogynist or not, and ii) whether the misogynist meme corresponds to shaming, stereotype, objectification, or violence classes. The first stage involved assigning binary labels to detect the presence or absence of misogyny and subclassification of misogynous memes. Figure~\ref{fig:MAMI_dataset_samples} presents a few sample memes from the MAMI dataset for manual understanding of its content and labeling structure.
The dataset supports two key tasks:

\begin{figure*}[!h] 
    \centering
    \includegraphics[width=1\textwidth]{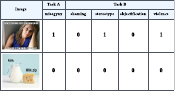} 
    \caption{Example from the MAMI dataset illustrating two tasks: Task A (binary: 1 = misogynistic, 0 = non-misogynistic) and Task B (multi-label: 1/0 for shaming, stereotype, objectification, violence).}
    \label{fig:MAMI_dataset_samples}
\end{figure*}

The details of the sub-tasks is given below-
\begin{itemize}
    \item \textbf{Sub-task A: Misogyny Detection} \\
    Classifies memes as \textit{Misogynous} or \textit{Non-Misogynous}.
    \item \textbf{Sub-task B: Misogyny Type Classification} \\
    Categorizes misogynous memes into one or more overlapping types:
    \begin{itemize}
        \item \textbf{shaming}: Criticism of women based on appearance or behavior.
        \item \textbf{Stereotype}: Imposition of traditional roles or fixed traits on women.
        \item \textbf{Objectification}: Treating women as objects.
        \item \textbf{violence}: Depictions or implications of physical or psychological violence against women.
    \end{itemize}
\end{itemize}

Table~\ref{tab:dataset_stat} summarizes the sample distribution for both the Memotion and the MAMI datasets. ``Avg. Length'' denotes the average sample length in words for each category.

\subsection{Configuration of VLMs}
\label{model_config}

In this study, we evaluate four open-source vision-language models (VLMs) for the meme classification task, as they are widely used in the image captioning and visual reasoning literature. The specific models used in our experiments are listed in Table~\ref{tab:vlm_version}, along with their implementation details.

\paragraph{LLAVA-1.6:}
LLAVA-1.6 (Large Language and Vision Assistant) (LV) \cite{liu2024llavanext} is a vision-language model optimized for multi-modal tasks. It extends LLAVA-1.5 \cite{liu2024improved} by incorporating improved instruction tuning and enhanced vision-language alignment, where CLIP \cite{radford2021learning} is used as the vision encoder and Vicuna \cite{vicuna2023} as the text encoder. 
\paragraph{Qwen2-VL:}
Qwen2-VL \cite{bai2024qwen2vl} is an advanced vision-language model, specifically designed to employ a mixed training regimen, both image and text. It utilized Vision Tr

\begin{table*}[!h]
\centering
\setkeys{Gin}{keepaspectratio}
\resizebox{0.8\textwidth}{!}{%
\begin{tabular}{|l|c|c|l|}
\hline
\textbf{VLM} & \textbf{Size} & \textbf{Multimodal Fusion} & \textbf{Version} \\
\hline
LLAVA-1.6 & 7B & Early fusion & \texttt{llava-hf/llava-v1.6-mistral-7b-hf} \\
\hline
Qwen2-VL & 7B & Early fusion & \texttt{Qwen/Qwen2-VL-7B-Instruct} \\
\hline
LLaMA-3.2-Vision & 11B & Multilevel fusion & \texttt{meta-llama/Llama-3.2-11B-Vision-Instruct} \\
\hline
InstructBLIP & 7B & Early fusion & \texttt{Salesforce/instructblip-vicuna-7b} \\
\hline
\end{tabular}
}
\caption{Selected VLMs and their corresponding versions used in our experiments.}
\label{tab:vlm_version}
\end{table*}

\subsubsection{Configuration of LLMs}
\label{llms_config}
To complement the vision-language understanding from VLMs, we use three pretrained language models—BERT (BR) \cite{devlin2019bert}, RoBERTa (RB) \cite{liu2019roberta}, and XLNet (XL) \cite{yang2019xlnet} —as classification backbones in our CoVExFiL pipeline (Exp 3). We have listed the particulars of the LLMs in Table~\ref{tab:llm_version}. These models are fine-tuned using the textual explanations generated by the VLMs, enabling a decoupled two-step classification process.

\paragraph{BERT (BR):}
BERT (Bidirectional Encoder Representations from Transformers) \cite{devlin2019bert} is a transformer-based model trained using masked language modeling. Its bidirectional attention makes it effective for capturing contextual dependencies in text. We use the base uncased version in our experiments.

\paragraph{RoBERTa (RB):}
RoBERTa \cite{liu2019roberta} improves upon BERT by training on larger datasets with dynamic masking and no next-sentence prediction. It is known for robust performance across many classification benchmarks.

\paragraph{XLNet (XL):}
XLNet \cite{yang2019xlnet} is an autoregressive model that incorporates permutation-based training to capture the bidirectional context. It overcomes some of the limitations of BERT, particularly in modeling word order and long-range dependencies.

\begin{table}[!h]
\centering
\setkeys{Gin}{keepaspectratio}
\resizebox{0.9\columnwidth}{\textheight}{%
\begin{tabular}{|l|c|r|}
\hline
\textbf{LLM} & \textbf{Size}  & \textbf{Version}  \\
\hline
BERT & 110M   & \texttt{bert-base-uncased} \\
\hline
RoBERTa & 125M  &  \texttt{roberta-base} \\
\hline
XLNet & 110M  & \texttt{xlnet-base-cased}  \\
\hline

\end{tabular}
}
\caption{Considered LLMs and their corresponding versions for our experiments.}
\label{tab:llm_version}
\end{table}

\subsubsection{Baselines}
The baseline models presented in Table~\ref{tab:performance} serve as reference points for evaluating performance on the Memotion and MAMI datasets. These models were selected based on their relevance to the respective tasks and their reported effectiveness in prior studies.
\begin{itemize}
    \item M2Seq2Seq-MLD ~\cite{zhang2023multitask}: A multi-task seq2seq model for multimodal sarcasm, sentiment, and emotion recognition. It captures intra- and inter-modality/context-task dynamics for SN and HM.

    \item MT-BERT+TextGCN ~\cite{kumari2024let}: Combines Multi-Task BERT with TextGCN for improved text classification via contextual and graph-based features.
    \item PBR (Pretraining-Based Representation) ~\cite{zhang2022srcb}: This model combines image features from CLIP with textual embeddings from BERT and UNITER, integrated through a late fusion strategy. The final predictions are made using an ensemble of XGBoost classifiers and refined through rule-based post-processing.
    \item BERT+ViT ~\cite{singh2023female}: This model combines a variant of the BERT language model, pretrained on hate-speech text data, with a Vision Transformer (ViT) serving as the visual encoder. 
\end{itemize}

\begin{table}[!h]
  \centering
  \setkeys{Gin}{keepaspectratio}
  \resizebox{\columnwidth}{!}{%
    \begin{tabular}{|l|l|r|l|}
      \hline
      \textbf{Category} & \textbf{Labels} & \textbf{SOTA} & \textbf{Model} \\
      \hline
      Sentiment   & Pos, Neu, Neg  & 38.30 & M2Seq2Seq-MLD \\ 
      \hline
      Humour      & [H, VF, F, NF]  & 49.09 & M2Seq2Seq-MLD \\ 
      \hline
      Sarcasm     & [VT, TM, S, G]  & 32.85 & MT-BERT+TextGCN \\ 
      \hline
      Offense     & [HF, VF, O, NO] & 34.38 & MT-BERT+TextGCN \\ 
      \hline
      Motivation  & [M, NM]         & 59.28 & MT-BERT+TextGCN \\ 
      \hline
      Misogyny & [Miso, Non-Miso] & 87.40 & BERT+ViT \\
      \hline
      Misogyny Type & [V, S, O, Sh] & 73.14 & PBR  \\
      \hline
    \end{tabular}%
  }
  \caption{Performance comparison across different tasks and metrics after removing baseline results.}
  \label{tab:performance}
\end{table}

\begin{figure*}[!h] 
    \centering
    \includegraphics[width=\textwidth]{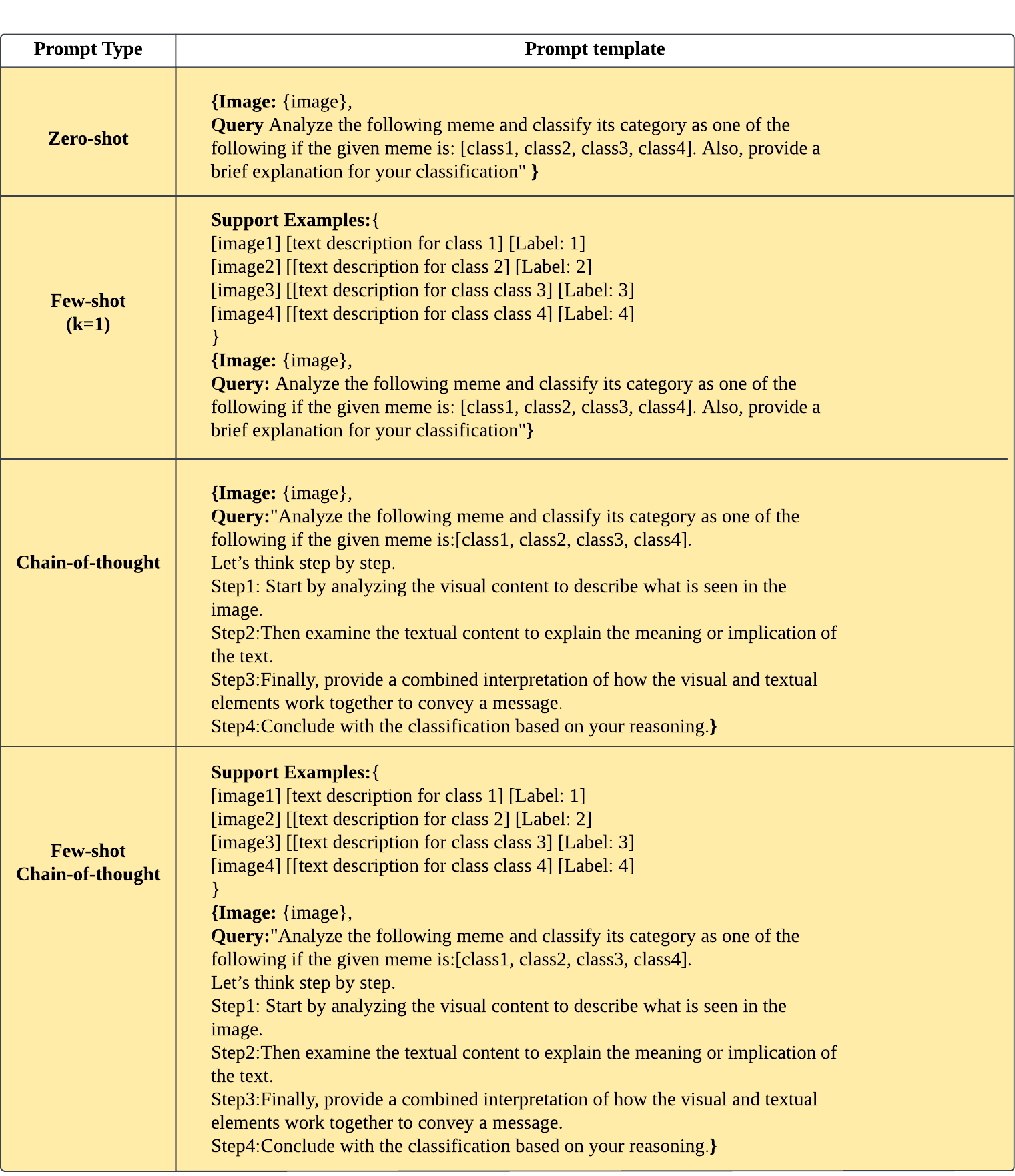} 
    \caption{Example Prompt Template for Experiment 1. Here, we have specified the prompts we used in ZS, ZSC, FS, and FSC.}
    \label{fig:prompt_template}
\end{figure*}

\begin{figure*}[!h] 
    \centering
    \includegraphics[width=\textwidth]{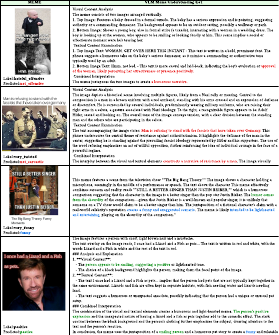} 
    \caption{Examples from the test set with their corresponding gold labels are shown to illustrate the VLM's understanding. Memes where the task was performed well are marked in \textcolor{green}{green}, those performed moderately well or relatable are marked in \textcolor{blue}{blue}, and those where the task was performed poorly are marked in \textcolor{red}{red}.}
    \label{fig:error_analysis_2}
\end{figure*}

\subsection{Configuration of Prompts}
\label{exp_1}
In Experiment 1, we explored a range of prompting strategies to evaluate their effectiveness in meme understanding. These included Zero-Shot (ZS), Zero-Shot Chain-of-Thought (ZSC), Few-Shot (FS), and Few-Shot Chain-of-Thought (FSC) prompting. The specific prompts used in each strategy, along with their corresponding examples and experimental settings, are illustrated in Figure~\ref{fig:prompt_template}. This figure presents detailed query formats and prompt structures for all four strategies, making it easier to understand the differences and design choices involved. Also for reproducibility we kept the temperature of the VLM very low (0.1)

\subsection{Configurations for Adapter-Based LoRA Fine-Tuning} \label{exp_2}
In this section, we reported the hyperparameters used for LoRA-based finetuning of VLMs. The list of hyperparameters are given in Table \ref{tab:hyperparams}.
For LLaMA, LLAVA, and QWEN-2-VL, we fine-tune LoRA adapters across all vision and language layers—including attention and MLP modules—to enable comprehensive adaptation of both image encoding and text generation, compensating for their lack of prior meme-specific instruction tuning. In contrast, we fine-tune only the q\_proj and v\_proj layers in InstructBLIP, targeting its core multimodal fusion mechanism to steer cross-modal alignment while minimizing additional memory and compute overhead.

\begin{table}[!h]
\centering
\setkeys{Gin}{keepaspectratio}
\resizebox{\columnwidth}{!}{%
\begin{tabular}{|l|l|}
\hline
\textbf{Hyperparameters} & \textbf{Values} \\
\hline
Rank             & 16 \\
\hline
Alpha ($\alpha$) & 32 \\
\hline
Learning rate    & 2e-4 \\
\hline
Optimizer        & \texttt{adamw\_8bit} \\
\hline
Epochs           & 2 \\
\hline
Layers fine-tuned & \texttt{vision\_layers}, \texttt{language\_layers}, \\
                  & \texttt{attention\_modules}, \texttt{mlp\_modules}, \\ &\texttt{q\_proj}, \texttt{v\_proj} \\
\hline
\end{tabular}%
}
\caption{Hyperparameters for LoRA fine-tuning of VLMs}
\label{tab:hyperparams}
\end{table}

\subsection{Performance Measures}
\paragraph{Evaluation metric:}  
Most meme classification tasks include significant class imbalance, as shown in Table ~\ref {tab:dataset_stat}. Due to this, we use the \textbf{weighted F1 score} as the primary evaluation metric to avoid bias toward majority classes. Unlike the macro F1 score, the weighted F1 accounts for class frequencies, providing a more unbiased assessment. We also evaluated the performance of the models in the average precision score.




\section{Error analysis}
\label{error_analysis}
The classification approach is insufficient for a thorough, comprehensive understanding of the ability of the VLMs. Hence, we further analyze the performance of the best models and prompting techniques using qualitative and quantitative methods.

\subsection{Quantitative analysis}
We evaluate the textual meme understanding generated by the VLMs in Experiment-3 using standard metrics—BLEU \cite{post2018call}, ROUGE \cite{lin2004rouge}, and BERTScore \cite{zhang2019bertscore}—following best practices in text generation evaluation, as noted by \citet{van2021human}. These metrics offer complementary insights: ROUGE captures surface-level similarity by measuring lexical overlap through n-grams and longest common subsequences. BLEU quantifies the precision of n-gram matches, making it useful for evaluating fluency and alignment with reference texts. BERTScore assesses semantic similarity using contextual BERT embeddings, enabling evaluation of deeper meaning beyond lexical matches.
We use these metrics because meme understanding requires evaluating both surface-level similarity and deeper semantic interpretation. While ROUGE and BLEU capture lexical overlap and fluency, BERTScore assesses semantic meaning, making it more suitable for our task.

To complete our quantitative analysis, we generated silver-standard textual explanations using the closed-source GPT-4 model \cite{achiam2023gpt}, since no meme dataset provides ground-truth explanations. All GPT-4 generations were performed in March 2025; as the model evolves, its outputs—and thus our evaluation scores—may change over time.

We compare the meme explanations generated by VLMs with those from GPT-4 using the metrics above. Table~\ref{tab:understanding_eval} presents the results of this comparison.

\begin{table*}
    \centering
    \setkeys{Gin}{keepaspectratio}
    \resizebox{0.9\textwidth}{!}{%
    \begin{tabular}{|l|l|l|l|l|l|l|l|l|l|l|l|}
    \hline
        \multirow{2}{*}{\textbf{Model}} & \multirow{2}{*}{\textbf{Prompting}} & \multicolumn{5}{c}{\textbf{Memotion}} & \multicolumn{5}{|c|}{\textbf{MAMI}} \\ \cline{3-12}
        ~ & ~ & \textbf{ROUGE-1} & \textbf{ROUGE-2} & \textbf{ROUGE-L} & \textbf{BLEU} & \textbf{BERT-Score} & \textbf{ROUGE-1} & \textbf{ROUGE-2} & \textbf{ROUGE-L} & \textbf{BLEU} & \textbf{BERT-Score} \\ \hline
        \multirow{4}{*}{LV} & ZS & 34.75 & 11.99 & 22.04 & 2.23 & 58.25 & 34.43 & 11.52 & 21.59 & 2.22 & 57.88 \\ \cline{2-12}
        ~ & ZSC & 50.97 & 20.99 & 26.58 & 13.43 & 64.23 & \textbf{48.91} & \textbf{19.14} & \textbf{25.77} & 10.89 & 62.59 \\ \cline{2-12}
        ~ & FS & 46.75 & 16.02 & 24.38 & 8.41 & 60.56 & 38.87 & 10.20 & 20.73 & 5.31 & 57.02 \\ \cline{2-12}
        ~ & FSC & 41.44 & 14.61 & 23.07 & 6.76 & 61.97 & 37.26 & 11.32 & 20.48 & 5.31 & 59.75 \\ \hline
        \multirow{4}{*}{QW} & ZS & 33.75 & 13.69 & 21.64 & 3.43 & 56.18 & 42.67 & 15.93 & 24.09 & 6.41 & 60.82 \\ \cline{2-12}
        ~ & ZSC & 50.99 & 20.92 & 26.75 & 13.32 & 64.48 & 41.71 & 9.52 & 19.43 & 4.84 & 57.36 \\ \cline{2-12}
        ~ & FS & 43.4 & 19.2 & 26.43 & 5.81 & 60.01 & 47.73 & 17.58 & 25.2 & 11.2 & \textbf{62.78} \\ \cline{2-12}
        ~ & FSC & 49.78 & 19.71 & 27.45 & 12.24 & \textbf{64.96} & 47.53 & 17.49 & 25.07 & 11.07 & 62.65 \\ \hline
        \multirow{2}{*}{LM} & ZS & 44.44 & 17.65 & 25.24 & 8.69 & 61.57 & 47.47 & 17.98 & 25.43 & 10.71 & 61.34 \\ \cline{2-12}
        ~ & ZSC & \textbf{51.25} & \textbf{21.23} & \textbf{26.84} & \textbf{13.67} & 64.57 & 48.00 & 18.86 & 24.32 & \textbf{11.24} & 61.69 \\ \hline
        \multirow{2}{*}{IB} & ZS & 7.17 & 5.22 & 6.95 & 1.1 & 40.33 & 7.34 & 5.08 & 7.01 & 0 & 40.85 \\ \cline{2-12}
        ~ & ZSC & 7.73 & 5.31 & 7.09 & 1.23 & 40.34 & 7.33 & 5.10 & 7.01 & 0 & 40.81 \\ \hline
    \end{tabular}}
    \caption{Evaluation performance of the considered VLMs versus GPT-4 generated understanding. The highest value of each metric is in \textbf{bold} (column-wise).}
    \label{tab:understanding_eval}
\end{table*}
We observe a correlation between better textual understanding and improved F1 scores in Table~\ref{tab:vlm_underst_llm_finetune_1}. Notably, the QWEN-2-VL model, which performs best in most Experiment-3 cases, also achieves the highest BERTScores across both datasets, indicating stronger semantic understanding.

\begin{figure*}[!h]
    \centering
\includegraphics[width=\linewidth]{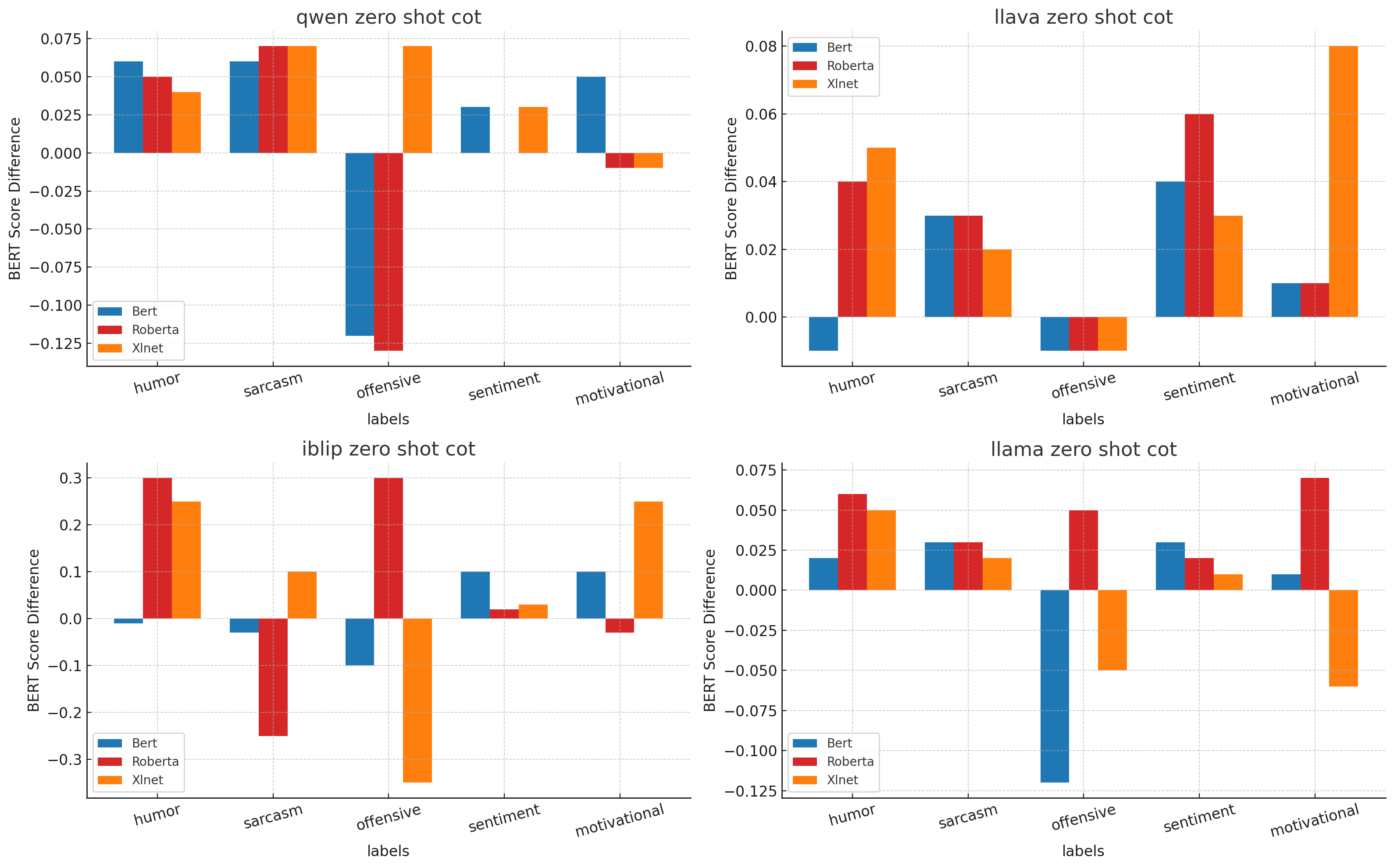} 
    \caption{BERT Score Difference between correctly classified vs misclassified samples}
    \label{fig:chart}
\end{figure*}

\subsection{BERTScore Difference Based Analysis}
We attempt to analyze the overall quality of the understanding generated by the four pre-trained VLMs under the CoT prompting in Experiment 3. For this purpose, we compare the quality of explanations across correctly and incorrectly classified samples. In order to quantitatively understand the rationale behind the performance of the classifications in Experiment 3. To do this, we computed the BERTScore difference between the two groups using GPT-4-generated silver labels as a reference. Figure~\ref{fig:error_analysis_2} shows the resulting differences across all tasks and models.

We selected the three-step CoT setting for this analysis because it achieved the highest average F1 score of 46.12 in Exp 3 . By focusing on CoT-prompted outputs from the four pre-trained VLMs, we aimed to determine whether better textual understanding, reflected through larger BERTScore differences, correlates with improved classification performance.

We based this analysis on the idea that higher BERTScore differences indicate more semantically meaningful explanations for correctly classified samples. These richer explanations likely guided the downstream LLMs during fine-tuning, resulting in accurate predictions. Conversely, smaller differences suggest that the explanations lacked sufficient semantic depth or clarity, which may have contributed to misclassification. In such cases, we observed that the explanations for misclassified samples often appeared lexically or semantically similar to those of correct predictions, highlighting the models' difficulty in capturing subtle linguistic and contextual cues. Figure~\ref{fig:error_analysis_2} presents the BERTScore differences across all tasks in the Memotion dataset. Our observations are as follows:

In these cases, the explanations for incorrect classifications were often lexically similar to those of correct classifications, suggesting that models struggled to distinguish subtle linguistic and contextual cues in SR and OF. The results indicate that SR and OF resulted in the lowest BERT Score differences, suggesting that the explanation quality for both correct and incorrect classifications is somehow similar for these categories. This observation highlights that fine-grained classification of SR and OF is challenging for AI models as it involves subtle linguistic and contextual cues. In contrast, SN and humor understanding resulted in relatively higher BERTScore differences, indicating that the VLMs produce a moderate BERT score for the memes they correctly classify. On the contrary, the model fails to comprehend the challenging memes, producing low BERT scores, leading to misclassification. We conclude from the observations that the VLMS are relatively better at identifying emotionally charged language or clear comedic elements in memes. However, motivational classification exhibited mixed performance, with some models relying on generic inspirational language without capturing deeper contextual meaning. Overall, the results highlight that while VLMs can effectively classify emotionally and socially sensitive aspects such as SN, they struggle with complex linguistic nuances like SR and OF.

\subsection{Qualitative analysis}
In this section, we examined misclassified samples, even by the best-performing VLMs in experiment 3. Figure~\ref{fig:error_analysis_2} shows a few such cases from the test set. The following are our findings:

We observed several interesting misclassification patterns across different meme understanding tasks, revealing limitations in the ability of the models to interpret visual and textual cues effectively. 

\begin{itemize}
    \item Example 1: In the offensiveness detection task, one meme was labeled as offensive but predicted as not\_offensive by the model. On closer inspection of the VLM's explanation, we find that it focuses on the childlike expressions and formal attire in the meme, failing to identify any hateful or offensive undertones. The elements highlighted by the model—marked in red—are entirely unrelated to offensive content, indicating a significant gap in detecting subtle hate cues.

    \item Example 2: In the sarcasm task, a meme was labeled as very\_twisted but misclassified as not\_sarcastic. The meme features a historical Nazi rally with a man defiantly refusing to perform the Nazi salute, surrounded by others, including Hitler. The accompanying text, “Man is refusing to stand with the fascists that have taken over Germany,” carries a sarcastic undertone, portraying irony and resistance. However, the model fails to detect this sarcasm, likely because the visual cue (Hitler's presence) is not referenced in the text, leading to confusion. The model does not sufficiently integrate contextual visual semantics with textual cues in this case.

    \item In the humor task, a meme with the true label very\_funny was only predicted as funny. The meme features a character from The Big Bang Theory with a humorous caption comparing his singing skills favorably to Justin Bieber's. While the model does recognize the meme as funny—which aligns with the VLM's explanation—it fails to differentiate between varying degrees of humor (i.e., funny vs. very\_funny). Penalizing this prediction as completely incorrect (i.e., score of 0) seems harsh, as the understanding is partially correct and only slightly off in granularity.

    \item Lastly, in the sentiment task, we found a case where the model correctly predicted the meme as positive, and its explanation also aligned well with both the visual and textual elements of the meme. The image features a person smiling while recalling a humorous pet story, which combines lighthearted visuals and an unexpected anecdote. The model’s interpretation here was accurate and demonstrates its potential when both modalities are clearly aligned.

\end{itemize}

Our analysis shows that current Vision-Language Models (VLMs) can understand basic aspects of memes but often miss deeper, context-driven cues. These models frequently rely too heavily on either the visual or textual modality, which causes them to overlook implicit offensiveness, miss sarcasm conveyed through visual irony, or misjudge the degree of humor. They also struggle to recognize culturally or historically significant visual elements when the accompanying text doesn’t make them explicit. Even when the models partially grasp the meme’s intent, strict evaluation metrics penalize them harshly, failing to account for close predictions. These findings highlight the need to develop models that align visual and textual information more effectively, reason contextually, and offer clearer explanations to handle the subtle and complex nature of meme understanding tasks.

\end{document}